\documentclass[conference]{IEEEtran}
\IEEEoverridecommandlockouts
\usepackage{cite}
\usepackage{amsmath,amssymb,amsfonts}
\usepackage{algorithmic}
\usepackage{subfig}
\usepackage{graphicx}
\usepackage{textcomp}
\usepackage{multicol}
\usepackage{multirow}
\usepackage{ctable}
\usepackage{xcolor}
\usepackage{comment}

\def\BibTeX{{\rm B\kern-.05em{\sc i\kern-.025em b}\kern-.08em
    T\kern-.1667em\lower.7ex\hbox{E}\kern-.125emX}}

\begin{document}

\newcommand{\etal}{\textit{et al.}}

\title{Reconfigurable Robots for Scaling Reef Restoration %\\
%\thanks{Reef Restoration \& Adaptation Program plus Matt's philanthropy sources of funding}
}
\author{\IEEEauthorblockN{Serena Mou, Dorian Tsai and Matthew Dunbabin}
\IEEEauthorblockA{\textit{Queensland Centre for Robotics} \\
{Queensland University of Technology}\\
Brisbane, Australia \\
serena.mou@hdr.qut.edu.au,~\{dy.tsai, m.dunbabin\}@qut.edu.au}

% \and
% \and
% \and
}

\maketitle

\begin{abstract}
% The Great Barrier Reef is under threat from climate change via bleaching. A variety of reef research projects are aimed at either transplanting coral larvae from healthy areas to not-healthy reefs, or growing and introducing heat-resistant corals to decimated reefs.  However, diving costs are expensive and to reach scale many current approaches are highly manual and do not scale. We thus propose to use robots to help out in certain areas to save cost and increase capacity for the production levels required to span the Great Barrier Reef. (1-2 sentence specifics)
% This paper focuses on the technical opportunities around employing 

% problem
Coral reefs are under increasing threat from the impacts of climate change. Whilst current restoration approaches are effective, they require significant human involvement and equipment, and have limited deployment scale. Harvesting wild coral spawn from mass spawning events, rearing them to the larval stage and releasing the larvae onto degraded reefs is an emerging solution for reef restoration known as coral reseeding.
% solution
This paper presents a reconfigurable autonomous surface vehicle system that can eliminate risky diving, cover greater areas with coral larvae, has a sensory suite for additional data measurement, and requires minimal non-technical expert training. 
A key feature is an on-board real-time benthic substrate classification model that predicts when to release larvae to increase settlement rate and ultimately, survivability.
% contributions/results
The presented robot design is reconfigurable, light weight, scalable, and easy to transport. Results from restoration deployments at Lizard Island demonstrate improved coral larvae release onto appropriate coral substrate, while also achieving 21.8 times more area coverage compared to manual methods.
% TODO: add in other concrete results
\end{abstract}

\begin{IEEEkeywords}
marine robotics, environmental robotics, deep learning for visual perception, coral reef restoration
\end{IEEEkeywords}

%%%%%%%%%%%%%%%%%%%%%%%%%%%%%%%%%%%%%%%%%%%%%%%%%%%%%%%%%%%%%%%%%%%%%%%%%%%%%%%%%%%%%%%%%
\section{Introduction}

% measurement collection and analysis, coordination and collaboration  between multiple heterogeneous data-sampling platforms, data fusion, sensor motion, target tracking,

% The Great Barrier Reef (GBR) is great, but the Great Barrier Reef is under threat from climate change. 
% Cite other coral reef initiatives (Cordrap, Neom, others in Bahamas, Philipines, etc).
Coral reefs are under extreme pressure from impacts linked to climate change~\cite{climate1999, bleaching2016, bleaching2017, decrease2012}. 
Coral reseeding is a restoration technique pioneered by Harrison \etal, where "slicks" of wild coral sperm and eggs released during mass spawning events are harvested from the ocean surface, fertilised, reared until the larval stage, and the larvae released onto target reefs~\cite{harrison2016coral}. This allows for the placement of the coral larvae onto appropriate substrates that greatly increase their chance of survival, and has been shown to successfully restore populations of coral~\cite{harrison2016coral, cruz2020movingcorals}. 
However, most reef restoration projects have been relatively small in spatial scale with the median size of restored reef of approximately 100 m$^2$\cite{review2020}. Additionally, current deployments involve manually intensive processes with significant diver and vessel costs~\cite{dunbabin2020uncrewedMaritimeSystems}. To perform restoration deployments at the scale of the Great Barrier Reef~\cite{intervention2017}, automation is essential~\cite{review2020, dunbabin2020uncrewedMaritimeSystems}. % should pull up more references for automation required?

% Describe the manual moving coral process. %  Process of this method - reference Peter Harrison
% However, the current practice is localised in its coverage, manually intensive, it requires divers which are expensive and have limited operation time and carrying capacity, and therefore, it does not scale well to the required operational levels. 
% As noted by Bostr{\"o}m-Einarsson \etal most reef restoration projects have been relatively small in spatial scale, with a median size of restored area of 100 m$^2$\cite{review2020}.

\begin{figure}[t!]
    \centering
    \includegraphics[width=1.0\columnwidth]{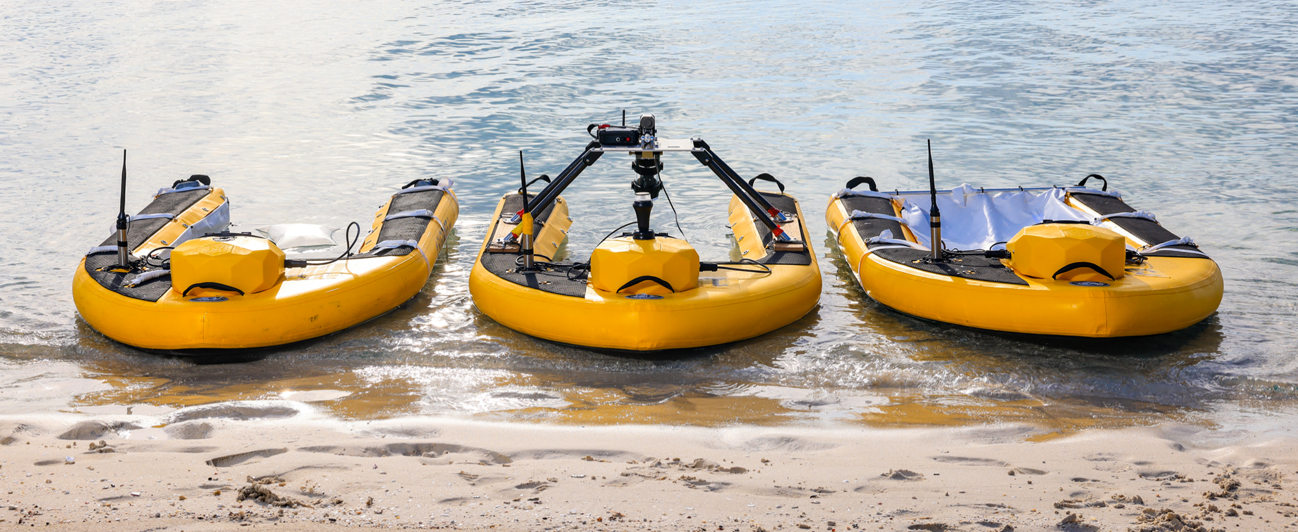}
    \caption{A sample fleet of modular FloatyBoats configured for capturing coral spawn (left), monitoring and 3D mapping reefs (middle), and deploying coral larvae (right).
    %Top view of Floatyboat during larval dispersal. Image courtesy of SCU
    }
    \label{fig:floatyboat_title}
\end{figure}

In this paper, we present \emph{FloatyBoat}, shown in Fig.~\ref{fig:floatyboat_title}, a novel Autonomous Surface Vehicle (ASV) which was developed to automatically and precisely deploy coral larvae across entire reefs.  
Our key contributions are the design of a reconfigurable ASV platform with an on-board automatic vision-based substrate classification algorithm that operates in real-time to control the larvae deployment pumps; all managed and deployed via an app. This allows for scaled up reef restoration that does not require trained divers or technical experts in the field. We demonstrated the utility of these contributions through restoration deployments at reefs off Lizard Island, the Great Barrier Reef, and in the Philippines. % add something about scaling up?

% \begin{figure}[htbp]
% \centerline{\includegraphics[width=\columnwidth]{figures/FleetFloatyBoat.png}}
% \caption{[placeholder] A fleet of FloatyBoats for coral larvae capture, monitoring and release, left to right, respectively. Ideally: FloatyBoat deploying larvae. Image credits}
% \label{fig:floatyboat_title}
% \end{figure}

% The rest of this paper is organised as follows. We first review related work in Section~\ref{sec:related}. In Section~\ref{sec:boat_design}, we discuss the FloatyBoat's design and the mechanisms/algorithms for coral larvae collection, dispersal and substrate classification. Next we present the field results from deployments on two reefs near Lizard Island in Section~\ref{sec:results}. % In Section~\ref{sec:scale}, we discuss the scalability of FloatyBoat. 
% Lastly, in Section~\ref{sec:conclusions}, we conclude the paper and explore avenues of future work.

%%%%%%%%%%%%%%%%%%%%%%%%%%%%%%%%%%%%%%%%%%%%%%%%%%%%%%%%%%%%%%%%%%%%%%%%%%%%%%%%%%%%%%%%%
\section{Related Work of Marine Robots}
\label{sec:related}

% limited scope: we are not discussing other methods of coral restoration. No space
% We consider previous work from other existing marine robots, and deep learning classification methods.
% In this section, we discuss related work for previous marine robots designed for reef restoration. % and the deep learning classification methods used for the substrate classification.

% Other methods of moving corals.
% \subsection{Marine Robots}

% here, we make the argument for needing modularity/multi-purpose robots
Existing underwater robots and surface vehicles are often highly-specialised for their application~\cite{bluefin2019,deepc2004,ice2020,cotsbot}. As a result of this specialisation, their usefulness for other tasks can be limited. Given the high costs of running a robot, a narrow usage case limits the usefulness and applicability of the robot for intensive tasks such as reef restoration. Reconfigurable robots are one approach to alleviate this issue.

Previously proposed for reef restoration, the Rangerbot Autonomous Underwater Vehicle (AUV) was developed for modularity and has been deployed for deep and precise coral larvae placement~\cite{dunbabin2020uncrewedMaritimeSystems}. 
% TODO: search for any other modular marine robot systems/check with Matt
Whilst suitable for deeper water deployments, the Rangerbot has a smaller larvae carrying capacity which ultimately lowers coverage. It is also unsuitable for harvesting coral slicks. A more suitably designed surface vehicle for coral reseeding offers a greater payload, endurance and multi-function opportunities.

%%%%%%%%%%%%%%%%%%%%%%%%%%%%%%%%%%%%%%%%%%%%%%%%%%%%%%%%%%%%%%%%%%%%%%%%%%%%%%%%%%%%%%%%%
\section{FloatyBoat: A Reconfigurable Autonomous Surface Vehicle}
\label{sec:boat_design}

% In this section, we describe the design considerations of the FloatyBoat Unmanned Surface Vehicle (USV), as well as the processes of coral larvae collection, reseeding and method for substrate classification.

% \begin{figure}
%   \centering
%   \includegraphics[width=0.7\columnwidth]{figures/floatyb.png}
%   %\includegraphics[scale=1.0]{figurefile}
%   \caption{System Components. A) Dwayne the Brain B) Net, Bladder or Camera C) Pump and hose D) Propellers E) Camera \textcolor{red}{3D model with labels available? Consider removing/replacing by adding labels to Fig. 3, 4.}}
%   \label{fig:boat_components}
% \end{figure}

\subsection{Design Considerations}

% modem, motor controllers, GPS, batteries, onboard computer
FloatyBoat was designed as a multi-purpose conservation tool to complement the RangerBot AUV for shallow reef restoration around the world~\cite{dunbabin2020uncrewedMaritimeSystems}. The key system components for coral slick collection and larvae deployment are shown in Fig.~\ref{fig:floatyboat_collect} and~\ref{fig:floatyboat_deliver} respectively.
%, with a comparison to divers and Rangerbot shown in Table~\ref{tbl:comp}. 
FloatyBoat ASV uses two thrusters for differential control, giving it an endurance of two hours at 0.75 m/s on a single battery. The sensing payload for reef restoration is typically configured with a camera and a small onboard computer with a GPU (NVIDIA Jetson Nano) for image processing, and GPS, compass and depth sounder for remote control or autonomous navigation. 
The controller is capable of following a path with less than 0.5 m cross track error, and can coordinate with other ASVs in swath formations such as a line and V-patterns.

The relatively low-cost of FloatyBoat stems from the use of high-performance micro-controllers and the ability to quickly reconfigure the robot without any structural changes. The U-shape of the boat was a deliberate decision to allow the additions of payload systems such as spawn collection nets (Section~\ref{subsec:design_collect}), larvae bladders (Section~\ref{subsec:design_disperse}), and retractable sensors (e.g. high-resolution cameras and sonars) for mapping and monitoring (Section~\ref{subsec:design_nav}). The ASV can be deflated and packed for transport, then re-inflated with a hand pump upon arrival. 
% The trade-off is FloatyBoat's ability to release larvae at close proximity to the substrate in deeper water, further discussed in Section~\ref{subsec:design_disperse}.

\subsection{Coral Larvae Collection}
\label{subsec:design_collect}
% Describe the process of collecting coral larvae.

\begin{figure}
    \centering
    \subfloat{\includegraphics[height=0.42\columnwidth, width=0.495\columnwidth]{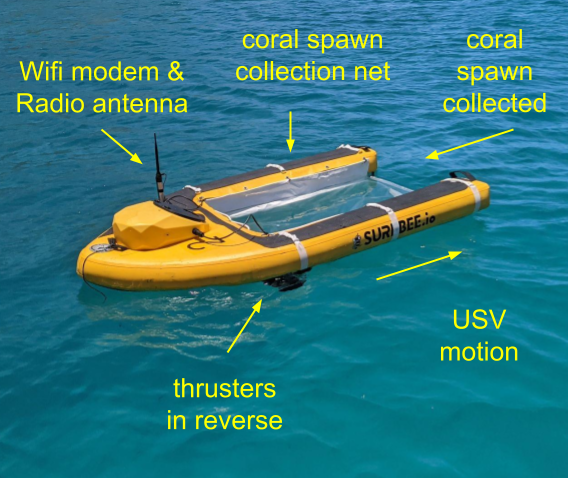}} \hfill
    \subfloat{\includegraphics[height=0.42\columnwidth, width=0.495\columnwidth]{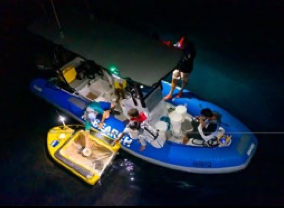}}
    \caption{FloatyBoat configured for larvae collection, driving backwards to scoop up coral spawn which floats at the surface (left), then scooped out into boats for collection (right).}
    \label{fig:floatyboat_collect}
\end{figure}

% The U-shape of the boat was a deliberate decision to allow the additions of the spawn collection net and the larvae bladder. When the surface vehicle is in collection mode, a net is attached inside the U and the boat is teleoperated mouth side forward. This allows the boat to gather up the spawn that is floating on the surface to be collected. When the surface vehicle is in dispersal mode, a large bladder nests inside the U-shape and a pump is added. This bladder is easily filled with the concentrated coral larvae. In this configuration, the boat is able to cover large areas of the reef with coral larvae. 
To collect coral spawn slicks, the FloatyBoat is configured with a fine mesh net as shown in Fig.~\ref{fig:floatyboat_collect}. In this mode, the controller is reversed to drive the ASV in the direction of the net opening, thus skimming the floating spawn from the water's surface into the net. 

\subsection{Coral Larvae Dispersal}
\label{subsec:design_disperse}
% Describe the physical process of pumping out the coral larvae.

\begin{figure}
    \centering
    \subfloat{\includegraphics[height=0.35\columnwidth, width=0.495\columnwidth]{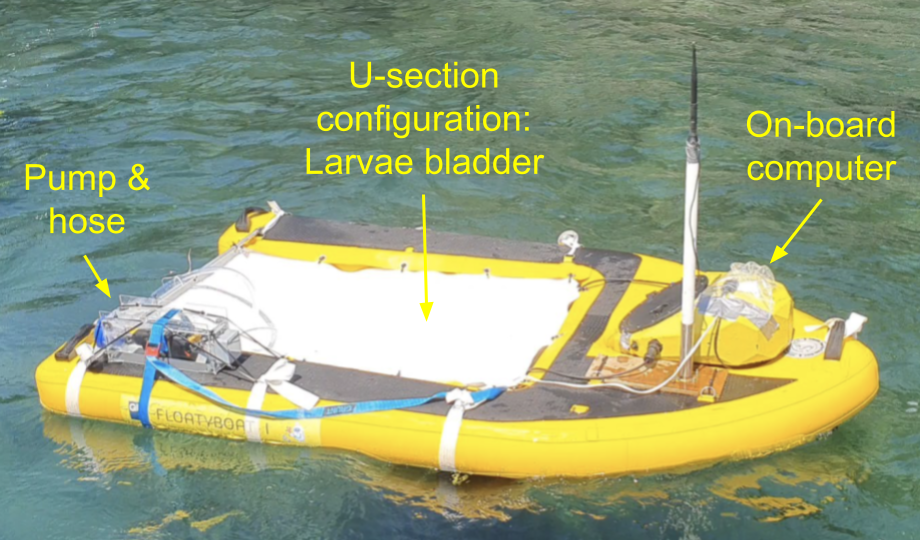}} \hfill
    \subfloat{\includegraphics[height=0.35\columnwidth, width=0.495\columnwidth]{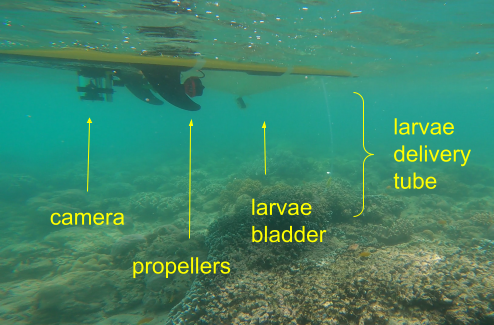}}
    \caption{FloatyBoat configured for larvae dispersal seen from above with the bladder (left), and below with the camera and larvae delivery tube visible (right).}
    \label{fig:floatyboat_deliver}
\end{figure}

% taken from \cite{dunbabin2020uncrewedMaritimeSystems}
For coral larvae dispersal, the FloatyBoat is configured with a 100L larval bladder, and a multi-pump with a multi-delivery hose system. Together, the system allows for larvae deployment rates of up to approximately 10,000 larvae per m$^2$ across the reef. The dispersal configuration is shown in Fig.~\ref{fig:floatyboat_deliver}. 

% Dorian: note sure if this is worth mentioning, since we do not yet reference it anywhere in the results section
\begin{comment}
\subsubsection{Deployment Modes}
The method of dispersal also plays an important role in larvae settling. 
Deployment modes of operation are (1) continuous, (2) pulsed with artificial intelligence (AI) decision support, and (3) constant pulse. Pulsing is an evolved behavior for deployment of larvae. Instead of continuously deploying, it moves 3-5 m then pulses to produce a low-velocity mushroom cloud of larvae to assist with settling. Then after a fixed time (determined based on seeding density requirements) the USV moves another 3-5 m along the transect and repeats. If the AI says it is suitable then it is pulsed, if not, then it moves another 3-5 m and tries again. 
\end{comment}

\subsubsection{Substrate Classification}
% why is coral substrate classification important
% how does this factor into the dispersal system
% how does the substrate classification work
Larvae dispersal is regulated by the coral substrate classification camera system. Coral larvae are more likely to settle near other hard corals and on solid rock; they are unlikely to settle on or near soft corals, sandy bottoms, algae or loose rubble substrate \cite{2020substrate}. Thus, a system that can decide whether the substrate below the ASV is suitable for restoration can improve settlement success and prevent wastage of precious larvae. 

%There was no need for localisation in the image or pixel-wise accuracy, classification task was chosen for its ease of labelling and low cost of running.
To determine the suitability of substrates, image classification was selected over pixel-wise segmentation -- primarily because localisation within the image was unnecessary and it allows for computationally cheaper models.
 % Note: large amounts of data not necessary
% We need a binary image classifier to say substrate or not. 
Convolutional Neural Networks (CNNs) are currently the state-of-the-art for image classification~\cite{krizhevsky2012imagenet}. MobileNetV3 is a CNN designed for mobile phone CPUs~\cite{howard2019searching}, and was thus ideal for lightweight mobile robotic platforms, such as the FloatyBoat. We adapted MobileNetV3 into a binary image classifier to predict whether or not a given image was either suitable or unsuitable substrate for coral recruitment. 

% Positive images are what our network should look for (substrate), while negative images tell our network what \textit{not} to look for (sand, soft corals, aglae, rubble). 
%, which is why transfer learning (cite) /few-shot learning (cite) is very important.
% could probably give data-set information in table format, but would be less compact!
% remove, if short on space
% For Heron Island, a Point Grey Grasshopper GS3-U3-23S6C-C camera was used with a Kowa 1-inch 8mm 5 MP c-mount lens to capture 590 images. Of these, 215 were negative, 199 were positive were used for training. For testing, 88 images positive, 88 images negative were used. For Lizard Island, an Alvium Allied Vision 1800U-1236C camera was used with the same lens. There were 635 positive, 629 negative images for training, and 200:200 positive:negative images for testing. \textcolor{red}{\textit{TODO: Explain how  this split decided?}}. 

\begin{ctable}[
    caption = {Benthic substrate classifier performance on image test data.},
    label = tbl:ML,
    doinside = \scriptsize,
    pos = t!,
    ]
    {ccccc}
    {}
    {\FL 
        \textbf{Train} & \textbf{Test} & \textbf{Accuracy} & \textbf{F1-Score} & \textbf{F1-Score} \\
         & &  \textbf{(\%)} &\textbf{(unsuitable)} & \textbf{(suitable)}\ML
        Lizard & Lizard & 89.74 & 0.9178 & 0.8629 \\

         & Heron & 89.97 & 0.9140 & 0.8779 \\
        \textbf{Average} &  & 89.86 & 0.9159 & 0.8704\\
        \FL
        Heron & Lizard & 98.54 & 0.9851 & 0.9856 \\
         & Heron & 99.61 & 0.9961 & 0.9961 \\ 
        \textbf{Average} &  & 99.08 & 0.9906 & 0.9909\\
        \FL
        Combined & Lizard & 99.95 & 0.9995 & 0.9995 \\
         & Heron & 98.98 & 0.9899 & 0.9896 \\
        \textbf{Average} &  &\textbf{99.47} & \textbf{0.9947} & \textbf{0.9946}\\
        \FL
 }
\end{ctable}

\begin{figure}[tb]
  \includegraphics[width=1.0\columnwidth]{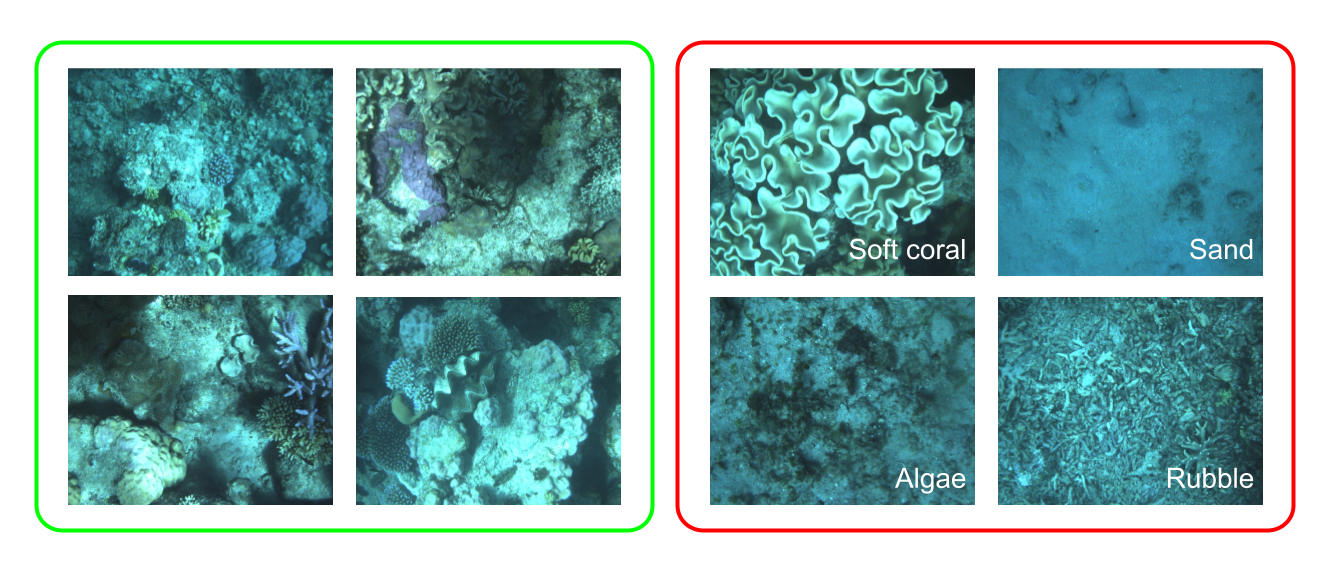}
  \caption{Suitable substrate (left) and unsuitable substrate (right) identified by the benthic substrate classification model.}
  \label{fig:subex}
\end{figure}

We obtained over 2,254 1920x1200 pix images from both Heron Island and Lizard Island to train and test our classifier. Images were labelled manually using expertise from marine scientists. To enable generalisation and robustness, data augmentations were applied during training. The images were randomly flipped and rotated, and small shifts to the colour-space were randomly applied. Table~\ref{tbl:ML} summarises the performance of the classifier. The images from each location were randomly split into balanced training, validation and test sets. The accuracy is shown along with the F1-score for each class. The F1-score incorporates the precision and recall of each class to provide a single more detailed metric for evaluation. Some typical examples of suitable (green) and unsuitable (red) substrates are shown in Fig.~\ref{fig:subex}. The highest performing model across both test sets was trained on both the Lizard and Heron data (Combined); however it should be noted that this was only by a small margin. 
% this makes sense: the model that is trained on both datasets/reefs, does the best on average across both reefs

%($F_{1}= 2·(\text{P} \times \text{R}) / (\text{P} + \text{R})$ where P = Precision and R = Recall ) 

\subsection{Coral Monitoring}
\label{subsec:design_nav}

\begin{figure}[tb]
    \centering
    \subfloat{\includegraphics[height=0.32\columnwidth, width=0.495\columnwidth]{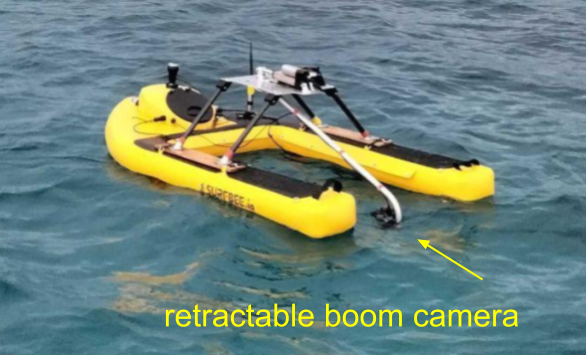}} \hfill
    \subfloat{\includegraphics[height=0.32\columnwidth, width=0.495\columnwidth]{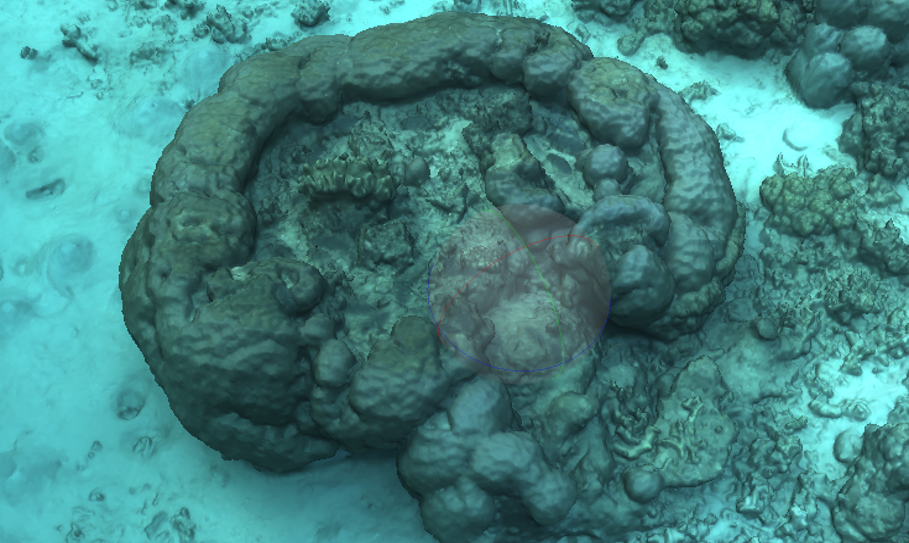}}
    \caption{FloatyBoat is configured with a retractable boom camera (left) for 3D reconstruction of coral reefs. 3D reconstructions of a large coral from Loomis reef (right) generated using imagery collected by the ASV.}
    \label{fig:camera}
\end{figure}

The coral monitoring configuration for FloatyBoat uses a retractable boom camera placed over the hull's U-section as in Fig.~\ref{fig:camera}. While the camera from the dispersal system was sufficient for navigation, control and substrate classification, the boom camera was designed to minimise self-shadow and the images were collected at a much higher resolution for 3D reconstruction. The ASV automatically traverses an area in a coverage pattern at a pre-selected track width that provides sufficient overlap for photogrammetry.

% \begin{figure}[tb]
%   \centering
%   \includegraphics[width=0.5\columnwidth]{figures/FloatyBoat_Pump_back.png}
%   \caption{FloatyBoat is configured with a boom camera for 3D reconstruction of coral reefs.}
%   \label{fig:camera}
% \end{figure}

\subsection{FloatyBoat Operation}
% Describe the FloatyBoat training app/GUI?

An essential aspect of using robots for large-scale deployments is that non-technical experts must be able to operate the systems. The ASV is tasked using an app with a graphical user interface (GUI) that focuses on high-level control, such as waypoint planning and payload configuration-dependent operations. A ``fuel gauge''-like feedback was implemented to show an estimate of the larvae remaining in the bladder, and an overlay of dots across the trajectory was used to summarise suitable substrate on the GUI. Up to seven ASVs can currently be controlled using the app. Live images from the ASV can be displayed, although bandwidth limitations during deployments currently prevent doing so in the field.

%%%%%%%%%%%%%%%%%%%%%%%%%%%%%%%%%%%%%%%%%%%%%%%%%%%%%%%%%%%%%%%%%%%%%%%%%%%%%%%%%%%%%%%%%
\section{Field Results}
\label{sec:results}

FloatyBoats were deployed across several reefs for coral restoration. On the Great Barrier Reef, two ASVs were used at Lizard Island (Loomis and Watson Reefs) in 2021, Heron Island Reef in 2020, as well as in the severely degraded reefs in the Philippines in 2021 and 2022. At each deployment, the collection, dispersal, mapping and training processes were refined.

\subsection{Coral Larvae Collection}

Larvae collection was successfully and efficiently accomplished using two FloatyBoats. With only 8 x 100 m transects, over 170L of larvae were collected during the 2021 Lizard Island spawning. The total collection time was $\sim$50 min with only two human operators. This time included re-positioning the ASVs three times due to being blown off station by wind. While no direct comparison to manually collecting coral spawn was made, under sparse slick conditions the ASVs provide much farther reach and maneuverability compared to a single crewed boat with many deck hands and buckets. % how often/common are sparse slick conditions?

\subsection{Coral Larvae Dispersal}

% % Pump test:
% A survivability test was conducted using competent larvae at the Lizard Island Research Station. 600 competent larvae were counted and split equally into two buckets (300 each) with approximately 5lt of seawater in each container. Larvae from one container were completely transferred into a separate container using the ZCLP at 30 rpm. The other (unpumped) larvae container was left as a control. The pumped larvae were immediately re-counted and all larvae were visually undamaged and competent. After one hour, both the pumped and control larvae were re-counted again with all larvae visibly healthy, then transferred into 2lt, clear containers.

% Two 5x5cm travertine settlement tiles were then placed into the container with the pumped larvae and also the container with the control larvae. After 24 hours, the tiles were counted for settled larvae with 46 (28+18) settlers on the control tiles and 50 (33+17) on the pumped larvae tiles.

% The results from the ZCLP trials showed that they had no observed impact on larvae survivorship under test conditions. The effects of larvae storage in the ASV bladder prior to pumping were not evaluated in this field campaign although will need to be in the future. The overall performance of the three pumps used were consistent, although there was one temporary failure on one pump due to a loose bolt that was later rectified. During the field deployments the delivery hose was a fixed length. However, in future deployments this should be adjustable using a winch or other mechanism to ensure the outlet is within cm of the seafloor.

\begin{figure}[tb]
  \includegraphics[width=\columnwidth]{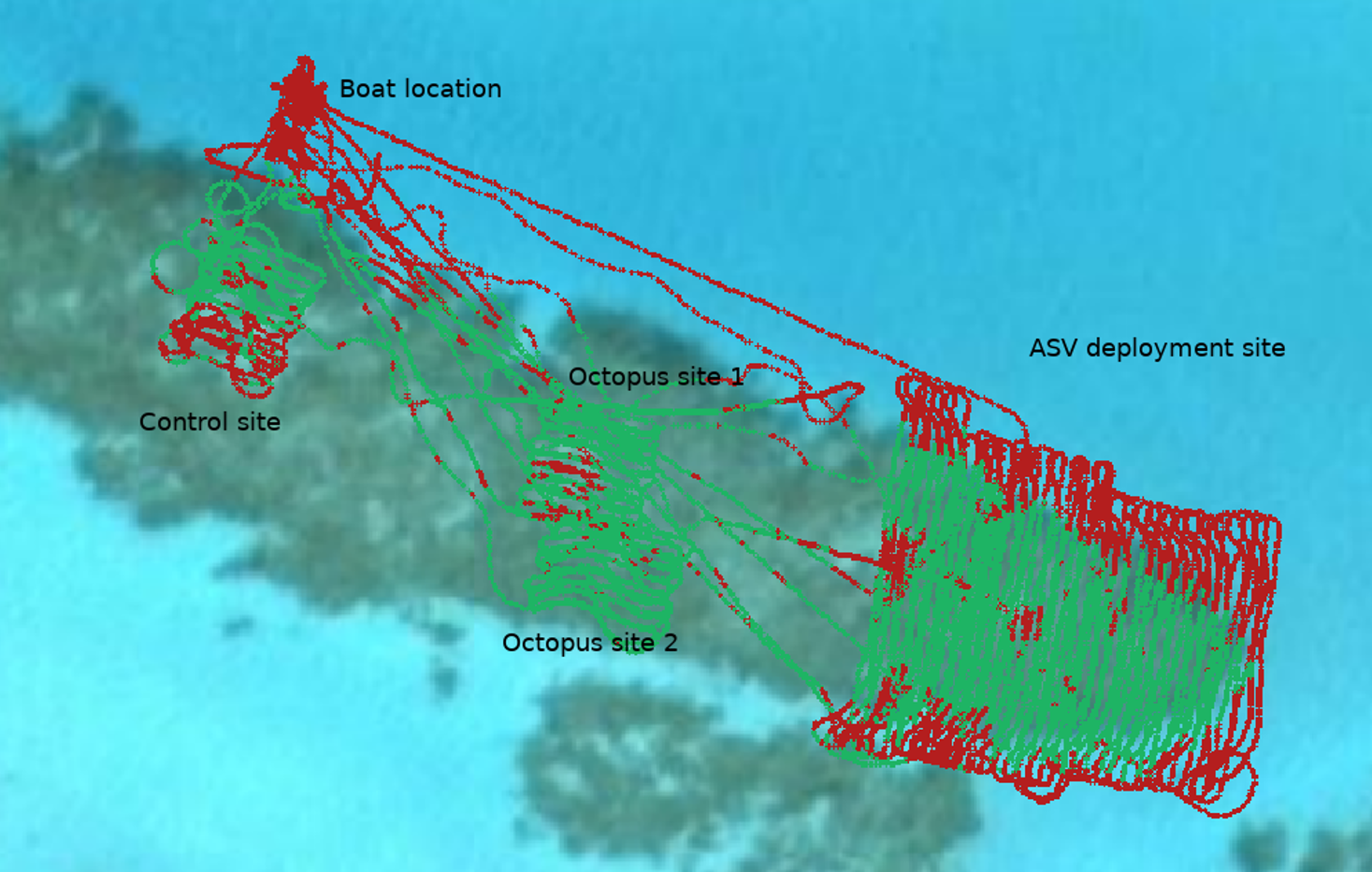}
  \caption{ASV trajectory over Loomis Reef, where the classifier regulated larvae dispersal. Larvae were released over suitable substrate (green) and retained over unsuitable substrate (red).}
  \label{fig:trajectory}
\end{figure}

The ASV trajectory during larvae dispersal at Loomis reef is shown in Fig.~\ref{fig:trajectory}. Green dots indicate suitable substrate according to the on-board substrate classification model, while red dots denote unsuitable substrate for larvae release. The two ``Octopus`` sites are manual coral dispersal deployments. The total manual deployments (across Watson and Loomis reefs) covered an area of $\sim$50 m$^2$. Conversely, the FloatyBoats operated mostly autonomously to release coral larvae over 200 m$^2$ at Watson's reef, and 890 m$^2$ at Loomis reef, achieving 21.8 times more coverage compared to manual deployments during the same time period.

% Table is the suitable substrate (TP), unsuitable substrate (TN), missed event (False negative), Wasted Larvae (False Positive), reframed in terms of the percentage of area covered by the ASV and the amount of wasted larvae.
The performance of the on-board substrate classifier in terms of the percentage of area covered by the ASV and the amount of larvae used is shown in Table~\ref{tbl:coverage}. The ground truth was obtained by manually labelling the images recorded during dispersal. The model was compared to a hypothetical case of constantly pumping larvae, whereby all larvae released over unsuitable substrate would be considered wasted. 
The on-board model correctly identified suitable substrate with a success rate of 98.8\% on Loomis and 98.7\% on Watson, with less than 1.13\% missed events, and less than 0.1\% wasted larvae. 
Using the classifier to regulate dispersal resulted in significantly less wasted larvae over sparse reefs (e.g. Loomis). While still wasting less larvae than the constant pump, the classifier was less impactful on dense coral-cover reefs (e.g. Watson).

% When the model correctly identified suitable substrate, a true positive was recorded. A true negative was recorded when the model correctly identified unsuitable substrate. Missed events, or false negatives, were recorded when the model incorrectly identified a suitable substrate as unsuitable. False positives were recorded when the model incorrectly identified unsuitable substrate as suitable, thereby wasting larvae. The on-board model was able to correctly identify suitable substrate with a success rate of 98.8\% on Loomis and 98.7\% on Watson, with less than 1.13\% missed events, and less than 0.1\% wasted larvae. 
% It is clear that using the classifier to regular dispersal results in significantly less wasted larvae over sparse reefs like Loomis, but is less impactful (though still useful) for densely-packed reefs like Watson.

% When the model correctly identified suitable substrate, a true positive was recorded. A true negative was recorded when the model correctly identified unsuitable substrate. Missed events, or false negatives, were recorded when the model incorrectly identified a suitable substrate as unsuitable. False positives were recorded when the model incorrectly identified unsuitable substrate as suitable, thereby wasting larvae. 

\begin{ctable}[
    caption = {Percentage of Area Covered by the ASV},
    label = tbl:coverage,
    doinside = \scriptsize,
    width = \columnwidth,
    pos = t!,
    ]
    {lcccc}
    {}
    {\FL  & 
        \begin{tabular}[l]{@{}l@{}}\textbf{Suitable}\\ \textbf{Substrate (\%)}  \end{tabular} & 
        \begin{tabular}[l]{@{}l@{}}\textbf{Unsuitable}\\ \textbf{Substrate (\%)}  \end{tabular} & 
        \begin{tabular}[l]{@{}l@{}}\textbf{Missed}\\ \textbf{Event (\%)}  \end{tabular} & 
        \begin{tabular}[l]{@{}l@{}}\textbf{Wasted}\\ \textbf{Larvae (\%)}  \end{tabular} \ML
        \textbf{Loomis Reef} & & & & \ML
        Ground truth & 46.85 & 53.15 & N/A & N/A \\
        Constant pump & 46.85 & N/A & N/A & 53.15 \\
        On-board model & 46.27 & 53.06 & \textbf{0.58} & \textbf{0.09} \ML
        \textbf{Watson's Reef} & & & & \ML
        Ground truth & 90.41 & 9.59 & N/A & N/A \\
        Constant pump & 90.41 & N/A & N/A & 9.59 \\
        On-board model & 89.28 & 9.49 & \textbf{1.13} & \textbf{0.10} \LL
 }
\end{ctable}

% \begin{figure}[tb]
%   \includegraphics[width=\columnwidth]{figures/image.png}
%   \caption{Table - convert to actual latex table}
%   \label{fig:classificationresults}
% \end{figure}
% \begin{figure}[htb]
%     \centering
%     \subfloat{\includegraphics[width=0.9\columnwidth]{figures/clam2-CLU.png}} \\
%     \subfloat{\includegraphics[width=0.9\columnwidth]{figures/clam5CLU.png}}
%     \caption{[placeholder] Amount of coral larvae used in a range of runs, to update with loomis and watson data}
%     \label{fig:larvae_used}
% \end{figure}

% \subsection{Coral Substrate Classification}

\subsection{Mapping}

Using the images collected from the FloatyBoat's camera, high-resolution 3D reconstructions of the reef systems can be generated. An example from Loomis Reef is shown in Fig.~\ref{fig:camera}. 

%Processing was performed offline. \textcolor{red}{How do we know these are accurate/good?}

\begin{comment}
\begin{figure}
    \centering
    \subfloat{\includegraphics[height=0.32\columnwidth, width=0.495\columnwidth]{figures/coral_reconstruction.png}} \hfill
    \subfloat{\includegraphics[height=0.32\columnwidth, width=0.495\columnwidth]{figures/coral_reconstruction_lizard.png}}
    \caption{Coral 3D reconstructions of a coral bommy (left), and Loomis reef (right) generated using imagery collected by the ASV.}
    \label{fig:reconstruction}
\end{figure}
\end{comment}

\subsection{Robot Transfer Training}

% Section to describe the current state of the FloatyBoat app and how it was received.

\begin{figure}
    \centering
    \subfloat{\includegraphics[height=0.32\columnwidth, width=0.495\columnwidth]{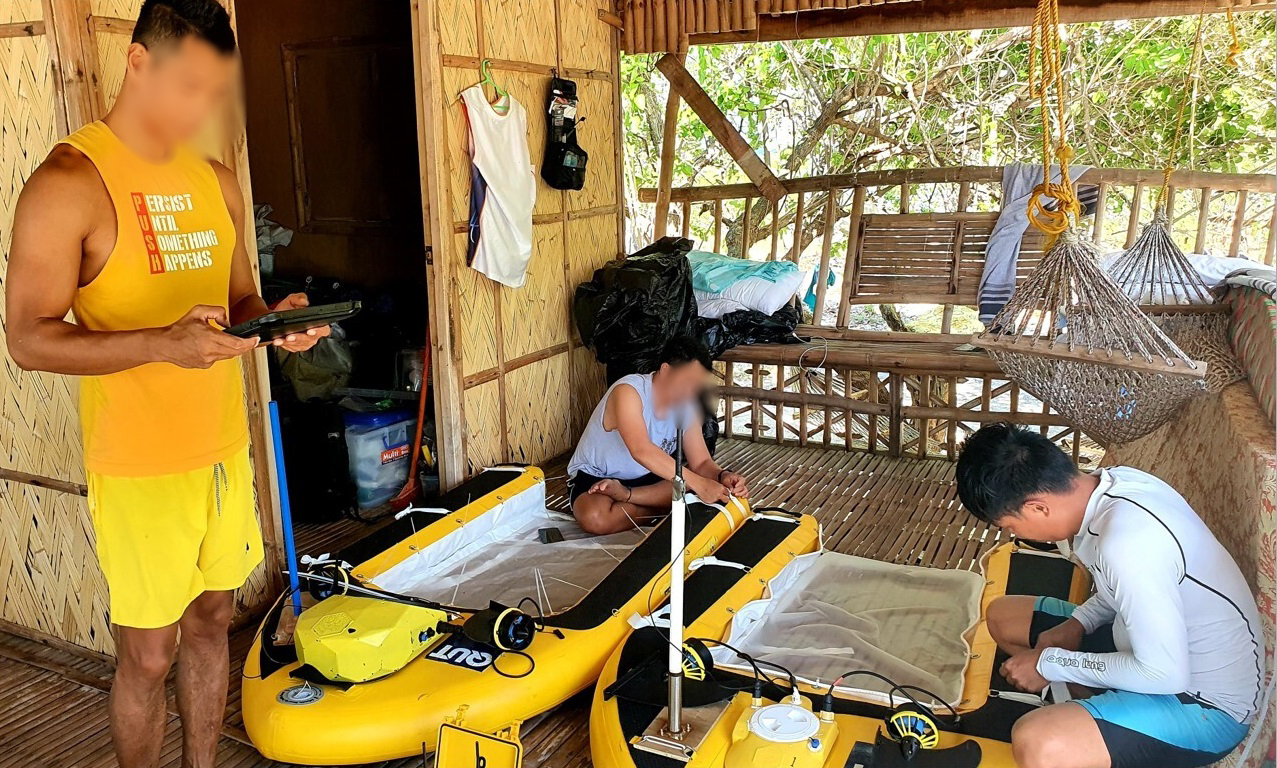}} \hfill
    \subfloat{\includegraphics[height=0.32\columnwidth, width=0.495\columnwidth]{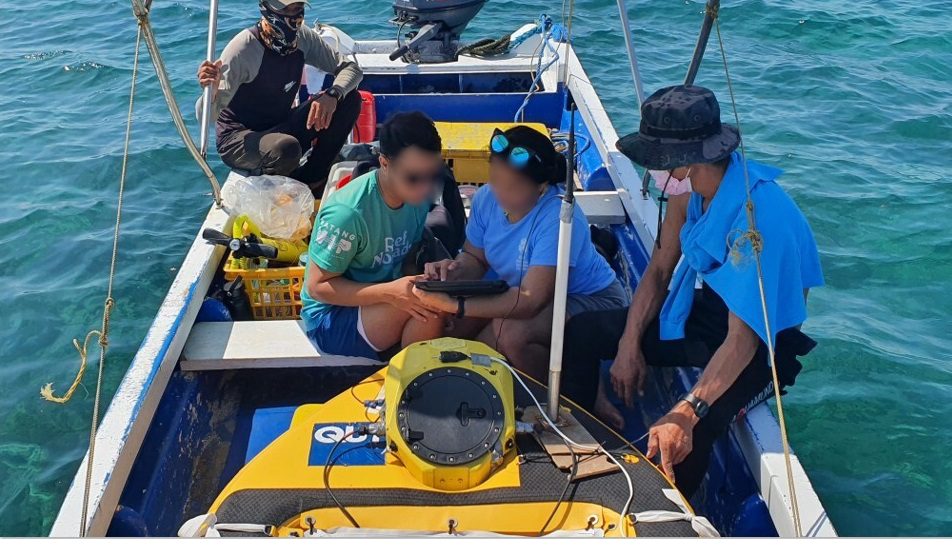}}
    \caption{ASVs being assembled by non-technical experts (left) guided only by the ASV app. (right) A single non-technical expert with only 10 minutes ASV training transfer trains others during the 2021 Philippines deployments.}
    \label{fig:training}
\end{figure}

In the Philippines deployment, FloatyBoats were successfully assembled by non-technical experts using only the ASV app. We employed transfer training, where we trained one non-technical expert in 10 minutes how to operate the ASV and this person then went on to train others. 
%Within only \textcolor{red}{30?} minutes, over \textcolor{red}{6?} transfer-trained persons were capable of operating FloatyBoat.
Fig.~\ref{fig:training} shows the successful self-guided assembly and transfer training by non-technical experts. These operators successfully operated the ASVs for spawn collection and larvae deployment at two deployment sites.

%\textcolor{red}{How did we verify they had successfully been trained?}
 
%%%%%%%%%%%%%%%%%%%%%%%%%%%%%%%%%%%%%%%%%%%%%%%%%%%%%%%%%%%%%%%%%%%%%%%%%%%%%%%%%%%%%%%%%
% \section{Scalability}
% \label{sec:scale}

% What we need to do to scale the operations up to GBR scale of millions of corals/year.
% Diver to robot cost comparison, coverage comparison, manufacturing, education/training app?

%%%%%%%%%%%%%%%%%%%%%%%%%%%%%%%%%%%%%%%%%%%%%%%%%%%%%%%%%%%%%%%%%%%%%%%%%%%%%%%%%%%%%%%%%
\section{Conclusions}
\label{sec:conclusions}

This paper has provided an overview of some design insights and operational approaches developed for ASVs to scale coral reef restoration. A reconfigurable, lightweight, easily transportable ASV was presented. 
Coral larvae dispersal was automatically regulated using an on-board real-time benthic substrate classification model, which conserved significant numbers of larvae, especially for ``sparse'' reef systems.
Multiple FloatyBoats were successfully used across several reefs with 21.8 times more reef coverage than manually deployed reef restoration projects.
Lastly, transfer training was demonstrated to promote operation of FloatyBoats for non-technical experts.
Combined, this work has demonstrated the role autonomous systems can have in scaling reef conservation.
%was done to help scale up reef restoration project operations and help preserve the reef.

% future work?

\section*{Acknowledgment}
This research was supported by a US-based
sponsor through the Friends of QUT in America Foundation. The authors thank Peter Harrison from the Southern Cross University and coral larvae reseeding collaborators through the Australian Centre for International Agricultural Research grant scheme.

\bibliographystyle{IEEEtran}
\bibliography{bibliography/coral.bib}

\end{document}